\documentclass[11pt,a4paper,leqno]{article}
\usepackage[hyperref]{pyBART}
\usepackage{times}
\usepackage{latexsym}
\usepackage{graphicx}
\usepackage{xytree}
\usepackage{tikz-dependency}
\usepackage{amsmath}
\usepackage{listings}
\usepackage{soul}
\definecolor{codegreen}{rgb}{0,0.6,0}
\definecolor{codegray}{rgb}{0.5,0.5,0.5}
\definecolor{codepurple}{rgb}{0.58,0,0.82}
\definecolor{backcolour}{rgb}{0.95,0.95,0.92} 
 
\lstdefinestyle{mystyle}{
    backgroundcolor=\color{backcolour},   
    commentstyle=\color{codegreen},
    keywordstyle=\color{magenta},
    numberstyle=\tiny\color{codegray},
    stringstyle=\color{codepurple},
    basicstyle=\ttfamily\footnotesize,
    breakatwhitespace=false,         
    breaklines=true,                 
    captionpos=b,                  
    keepspaces=true,                 
    numbers=left,                    
    numbersep=5pt,                  
    showspaces=false,                
    showstringspaces=false,
    showtabs=false,                  
    tabsize=2
}
 
\usepackage{url}
\graphicspath{ {./images/} }

\aclfinalcopy

\newcommand{\rep}{\textsc{BART}~}

\title{pyBART: Evidence-based Syntactic Transformations for IE}

\author{Aryeh Tiktinsky \hspace{1.5em} Yoav Goldberg \hspace{1.5em} Reut Tsarfaty \\
Allen Institute for AI, Tel Aviv, Israel \\
Bar Ilan University, Ramat-Gan, Israel \\
\texttt{\{aryeht,yoavg,reutt\}@allenai.org}}

\date{}

\begin{document}
\maketitle
\abovedisplayskip=0pt
\belowdisplayskip=0pt
\abovedisplayshortskip=0pt
\belowdisplayshortskip=0pt
\begin{abstract}
Syntactic dependencies can be predicted with high accuracy, and are useful for both machine-learned and pattern-based information extraction tasks. However, their utility can be improved. These syntactic dependencies are designed to accurately reflect syntactic relations, and they do not make semantic relations explicit. Therefore, these  representations lack many explicit connections between content words, that would be useful for downstream applications. 
Proposals like English Enhanced UD improve the situation by extending universal dependency trees with additional explicit arcs. However, they are not available to Python users, and are also limited in coverage. We introduce 
a broad-coverage, data-driven and linguistically sound set of transformations, that makes event-structure and many  lexical relations explicit.
We present pyBART, an easy-to-use  open-source Python library for converting English UD trees either to Enhanced UD graphs or to our representation. The library can work as a standalone package or be integrated within a spaCy NLP pipeline.
When evaluated in a pattern-based relation extraction scenario, our representation results in higher extraction scores than Enhanced UD, while requiring fewer patterns.
\end{abstract}

\section{Introduction}

\begin{figure*}
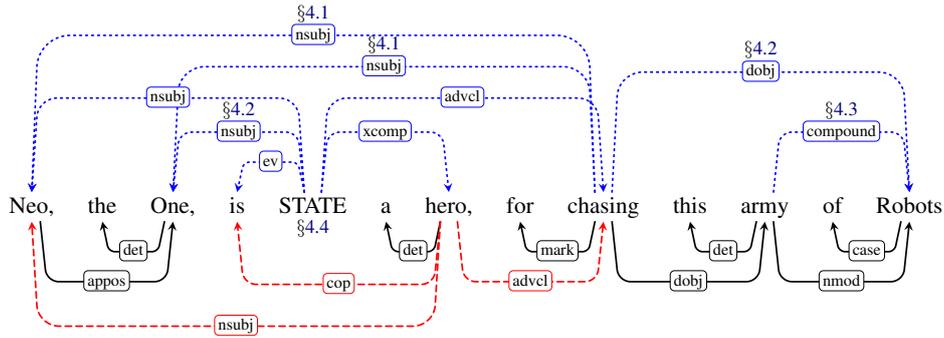

\centering
\scalebox{0.8}{
    \begin{dependency}[theme = default, label style={draw=blue}, edge style={thick, blue, dotted}]
       \begin{deptext}[column sep=1em]
          Neo, \& the \& One, \& is \& STATE \& a \& hero, \& for \& chasing \& this \& army \& of \& Robots \\
       \end{deptext}
       
       \depedge[edge below, label style={draw=black}, edge style={black, solid}]{1}{3}{appos}
       \depedge[edge below, label style={draw=black}, edge style={black, solid}]{3}{2}{det}
       \depedge[edge below, label style={draw=black}, edge style={black, solid}]{7}{6}{det}
       \depedge[edge below, label style={draw=black}, edge style={black, solid}]{11}{10}{det}
       \depedge[edge below, label style={draw=black}, edge style={black, solid}]{13}{12}{case}
       \depedge[edge below, label style={draw=black}, edge style={black, solid}]{9}{11}{dobj}
       \depedge[edge below, label style={draw=black}, edge style={black, solid}]{11}{13}{nmod}
       \depedge[edge below, label style={draw=black}, edge style={black, solid}]{9}{8}{mark}
       \depedge[edge below, edge height=6ex, label style={draw=red}, edge style={red,densely dashed}]{7}{4}{cop}
       \depedge[edge below, edge height=10ex, label style={draw=red}, edge style={red,densely dashed}]{7}{1}{nsubj}
       \depedge[edge below, label style={draw=red}, edge style={red,densely dashed}]{7}{9}{advcl}
       \depedge[edge height=12ex]{9}{3}{nsubj}
       \storelabelnode\secondlab
       \depedge[edge height=9ex]{5}{1}{nsubj}
       \depedge{5}{3}{nsubj}
       \storelabelnode\thirdlab
       \depedge{5}{4}{ev}
       \depedge{5}{7}{xcomp}
       \depedge[edge height=9ex]{5}{9}{advcl}
       \depedge[edge height=15ex]{9}{1}{nsubj}
       \storelabelnode\firstlab
       \depedge{9}{13}{dobj}
       \storelabelnode\forthlab
       \depedge{11}{13}{compound}
       \storelabelnode\fifthlab
       \node (silly1) [above of = \firstlab, yshift=-0.65cm, font=\small] {\S\ref{sec:nested}};
       \node (silly2) [above of = \secondlab, yshift=-0.65cm, font=\small] {\S\ref{sec:nested}};
       \node (silly3) [above of = \thirdlab, yshift=-0.65cm, font=\small] {\S\ref{sec:parallel}};
       \node (silly4) [above of = \forthlab, yshift=-0.65cm, font=\small] {\S\ref{sec:parallel}};
       \node (silly5) [above of = \fifthlab, yshift=-0.65cm, font=\small] {\S\ref{sec:alternation}};
       \node (silly6) [below of = \wordref{1}{5}, yshift=0.65cm, font=\small] {\S\ref{sec:state}};
    \end{dependency}}
    \caption{Representation of \emph{Neo, the One, is a hero, for chasing this army of Robots}. The arcs above the sentence are \rep additions. The ones below are EUD. Red arcs are removed in \rep while black are retained.}
    \label{fig:3}
\end{figure*}

Owing to neural-based advances in parsing technology, NLP researchers and practitioners  can now accurately produce  syntactically-annotated corpora at scale. However, the use and empirical benefits of the dependency structures themselves remain limited. Basic syntactic dependencies encode the functional connections between words but lack many functional and semantic relations that exist between the content words in the sentence. 
Moreover, the use of strictly-syntactic relations results in structural diversity, undermining the efforts to effectively extract coherent semantic information from the resulting structures.

Thus, human practitioners and applications that ``consume'' these  syntactic trees are  required to devote substantial efforts to  processing  the trees in order to identify and extract the information needed for downstream applications, such as information and relation extraction (IE). Meanwhile, semantic representations \cite{banarescu2013abstract, palmer2010semantic, abend2013universal, oepen-etal-2014-semeval} are harder to predict with sufficient accuracy, calling for a middle ground.

Indeed, \citet{de2008stanford} introduced \emph{collapsed} and {\em propagated} dependencies, in an attempt to make some semantic-like relations more apparent. The Universal Dependencies (UD) project\footnote{\url{universaldepdenencies.org}} similarly embraces the concept of \emph{Enhanced Dependencies} \cite{nivre-etal-2018-enhancing}),  adding explicit relations that are otherwise left implicit.  
\citet{schuster2016enhanced} provide further enhancements targeted  specifically at English (Enhanced UD).\footnote{In this paper we do not distinguish between the Universal Enhanced UD and \citet{schuster2016enhanced}'s {Enhanced++} English UD. We refer to their union on English as Enhanced UD.} \citet{candito-etal-2017-enhanced}  suggest  further enhancements to address  diathesis alternations.\footnote{Efforts such as PropS \cite{stanovsky2016getting} and PredPatt \cite{white-EtAl:2016:EMNLP2016}, share our motivation of extracting  predicate-argument structures from treebank-trainable trees, though  outside of the UD framework. Efforts such as KNext \cite{knext} automatically extract logic-based forms by converting treebank-trainable trees, for consumption by further processing. HLF \cite{hlf}, DepLambda \cite{deplambda} and UDepLambda \cite{udeplambda} attempt to provide
a formal semantic representation by converting dependency structures to  logical forms. While they share a high-level goal with ours --- exposing functional relations in a sentence in a unified way --- their end result, logical forms, is substantially different from pyBART structures. While providing substantial benefits for semantic parsing applications, logical forms are  less readable for non-experts than  labeled relations between content words. As these efforts rely on dependency trees as a backbone, they could potentially benefit from pyBART's focus on syntactic enhancements on top of (E)UD.}

In this work we continue this line of thought, and take it a step further. We present pyBART, an easy-to-use Python library   which converts English UD trees to a new representation that subsumes the English Enhanced UD representation and substantially extends it. We designed the representation to be linguistically sound and automatically recoverable from the syntactic structure, while exposing the kinds of relations required by IE applications.
Some of these modifications are illustrated in Figure \ref{fig:3}.\footnote{Some preserved UD relations are omitted for readability.} 
We aim  to  make event structure explicit, and cover as many linguistically plausible phenomena as possible.
We term our representation \rep (The BIU-AI2 Representation Transformation).

To assess the benefits of \rep with respect to UD and other enhancements, 
we compare them in the context of a pattern-based relation extraction task, and demonstrate that \rep achieves higher $F_1$ scores while requiring fewer patterns.

The python conversion library, \texttt{pyBART}, integrates with the spaCy\footnote{\url{https://spacy.io}} library, and is available under an open-source Apache license. A web-based demo for experimenting with the converter is also available. \url{https://allenai.github.io/pybart/}.

\section{The \rep Representation}
We aim to provide a representation that will be useful for downstream NLP tasks, while retaining the following key  properties. The proposal has to be {\bf (i) based on  syntactic structure} and {\bf (ii) useful for information seeking applications}. As a consequence of (ii), we also want it to {\bf  (iii) make event structure explicit} and {\bf (iv) allow favoring recall over precision}.

Being \textbf{based on syntax} as the backbone would allow us to capitalize on independent advances in syntactic parsing, and on its relative domain independence. We want our representation to be not only accurate but also \textbf{useful for information seeking applications}. This suggests a concrete methodology (\S\ref{sec:method}) and evaluation criteria (\S\ref{sec:eval}): we choose which relations to focus on based on concrete cases attested in relation extraction and QA-corpora, and evaluate the proposal based on the usefulness in a relation extraction task.

In general, information-seeking applications favor \textbf{making events explicit}. Current syntactic representations prefer to assign syntactic heads as root predicates, rather than actual eventive verb. In contrast, we aim to center our representation around the main event predicate in the sentence, while indicating event properties such as aspectuality (\emph{Sam \underline{started} walking}) or evidentiality (\emph{Sam \underline{seems} to like them}) as modifiers of rather than heads. To do this in a consistent manner, we introduce a new node of type STATE for copular sentences,  making their  event structure parallel to those containing finite eventive verbs  (\S\ref{sec:state})

Finally, downstream users may prefer to \textbf{favor recall over precision} in some cases. To allow for this, we depart from previous efforts that refrain from providing any uncertain information. We chose to explicitly expose some relations which we believe to be useful but judge to be uncertain, while clearly marking their uncertainty in the output. This allows users to experiment with the different cases and assess the reliability of the specific constructions in their own application domain. We introduce two uncertainty marking mechanisms, discussed in \S\ref{sec:unc}.

\subsection{Data-driven Methodology}
\label{sec:method}
Our departure point is the English EUD representation \cite{schuster2016enhanced} and related efforts discussed above, which we seek to extend in a way which is useful to information seeking applications. To identify relevant constructions that are not covered by current representations, we use a data-driven process. We consider concrete relations that are expressed in annotated task-based corpora: a relation extraction dataset (ACE05, \cite{walker2006ace}), which annotates relations and events, and a QA-SRL dataset \cite{he-etal-2015-question} which connects predicates to sentence segments that are perceived by people as their (possibly implied) arguments.
For each of these corpora, we consider the dependency paths between the annotated elements, looking for cases where a direct relation in the corpus corresponds to an indirect dependency path in the syntactic graph. We identify recurring cases that we think can be shortened, and which can be justified linguistically and empirically. We then come up with proposed enhancements and modifications, and verify them empirically against a larger corpus by extracting cases that match the corresponding patterns and browsing the results. 

\subsection{Formal Structure} \label{formal}
As is common in dependency-based representations, \rep structures are labeled, directed multi-graphs whose nodes are the words of a sentence, and the labeled edges indicate the relations between them.  Some constructions add additional nodes, such as copy-nodes \cite{schuster2016enhanced} and STATE nodes (\S\ref{sec:state}).

An innovative aspect of our approach is that each edge is associated with additional information beyond its dependency label. This information is structured as follows:

\noindent{\bf \textsc{Src}}: a field indicating the origin of this edge---either ``UD'' for the original dependency edges, or a pair indicating the type and sub-type of the construction that resulted in the \rep edge (e.g., \{\textsc{Src=}(conj,and)\} or \{\textsc{Src=}(adv,while)\}).

\noindent{\bf \textsc{Unc}, \textsc{Alt}}: optional fields indicating uncertainty, described below.

\subsection{Embracing uncertainty}
\label{sec:unc}
Some syntactic constructions are ambiguous with respect to the ability to propagate information through them. Rather than giving up on all ambiguous constructions, we opted to generate the edges and mark them with an \textsc{Unc=True} flag, deferring the decision regarding the validity of the edge to the user:
\\[0.5em]
\scalebox{0.8}{
    \begin{dependency}[theme = default, label style={draw=blue}, edge style={thick, blue, dotted}]
       \begin{deptext}[column sep=0.2cm]
          She \& acted, \& trusting \& her \& instincts \\
       \end{deptext}
       \depedge[edge height=2ex]{3}{1}{nsubj \{\textsc{Unc}\}}
       \depedge[edge height=2ex, edge below, label style={draw=black}, edge style={black, solid}]{2}{3}{dep}
       \depedge[edge height=2ex, edge below, label style={draw=black}, edge style={black, solid}]{2}{1}{nsubj}
    \end{dependency}}

In some cases, we can identify that one of two options is possible, but cannot determine which. In these cases we report both edges, but mark them explicitly as alternatives to each other. This is achieved with an \textsc{Alt=X} field on both edges, with \textsc{X} being a number indicating the pair.\\[0.5em]
\scalebox{0.8}{
    \begin{dependency}[theme = default, label style={draw=blue}, edge style={thick, blue, dotted}]
       \begin{deptext}[column sep=-0.1em]
          You \& saw \& me \& while \& driving, \& Sue \& saw \& Sam \& after \& returning \& \\
       \end{deptext}
       \depedge[edge height=3ex]{5}{3}{nsubj\{\textsc{Alt=0}\}} 
       \depedge[edge height=5.5ex]{5}{1}{nsubj\{\textsc{Alt=0}\}}
       \depedge[edge height=3ex]{10}{8}{nsubj\{\textsc{Alt=1}\}} 
       \depedge[edge height=5.5ex]{10}{6}{nsubj\{\textsc{Alt=1}\}}
    \end{dependency}}

\begin{figure}[t]
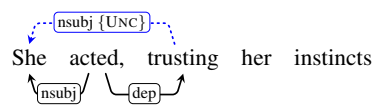

\begin{lstlisting}[language=Python]
# Load a UD-based english model
nlp = spacy.load("en_ud_model")

# Add BART converter to spaCy's pipeline
from pybart.api import converter
converter = converter( ... )
nlp.add_pipe(converter, name="BART")

# Test the new converter component
doc = nlp("He saw me while driving")
me_token = doc[2]
for par_tok in me_token._.parent_list:
    print(par_tok)

# Output:
{'head': 2, 'rel':'dobj', 'src':'UD'}
{'head': 5, 'rel': 'nsubj',
  'src':('advcl','while'), 'alt':'0'}
\end{lstlisting}
\caption{Usage example of pyBART's spaCy-pipeline component.}
\label{fig:5}
\end{figure}

\section{Python code and Web-demo}
\label{sec:software}

The py\rep library provides a Python converter from English UD trees to BART.
py\rep subsumes the enhancements of the EUD Java implementation provided in Stanford Core-NLP,\footnote{\url{https://nlp.stanford.edu/software/stanford-dependencies.html}} and extends them as described in \S\ref{structures}. While pyBART's default performs all enhancements, it can be configured to follow a more selective behavior. py\rep has two modes: (1) a converter from CoNLLU-formatted UD trees to CoNLLU-formatted \rep structures;\footnote{The extra edge information is linearized into the dependency label after a `@` separator.} and (2) a spaCy \cite{spacy2} pipeline component.\footnote{This requires a spaCy model trained to produce UD trees, which we provide.} After registering py\rep as a spaCy pipeline, tokens on the analyzed document will have a \texttt{.\_.parent\_list} field, containing the list of parents of the token in the \rep structure. Each item is a dictionary specifying---in addition to the parent-token id and dependency label---also the extra information described in \S\ref{formal}. See Figure \ref{fig:5} for an illustration of the API. 

A web-based demo that parses sentences into both EUD and \rep graphs, visualizes them, and compares their outputs, is also provided.\footnote{The dependency graph visualization component uses the TextAnnotationGraphs (TAG)  library \cite{TAG-2018}.} 

\section{Coverage of Linguistic Phenomena} \label{structures}

\rep conversion consists of four conceptual changes from basic UD. The first type propagates shared arguments between  predicates in
{\bf nested structures}. 
The second type  shares arguments between {\bf parallel structures}.
The third type attempts to unify {\bf syntactic alternations} to reduce diversity, making structures that carry similar meaning also similar in structure. Finally, the forth type is designed to make {\bf event structure} explicit  in the syntactic representation, 
allowing finite verbs that indicate event properties to act as event modifiers rather than root predicates.  In accordance with that, we further introduce a new STATE node, that acts as the main predicate node for {\em stative} (copular, verb-less) sentences.\vspace{-0.15em}

\subsection{Nested Structures}
\label{sec:nested}
Our first type of conversions propagates an external core argument to be explicitly linked as the subject of a subordinate clause.

\noindent{\bf Complement control:}
The various EUD representations explicitly indicate the external subjects of \emph{xcomp} clauses containing a \emph{to} marker. We embrace this choice and extend it to cover also clauses without a \emph{to} marker, including imperative clauses and clauses with controlled gerunds.\vspace{0.5em}
\begin{equation}
\label{complement-control1}
\begin{gathered}
\scalebox{0.8}{
    \begin{dependency}[theme = default, label style={draw=blue}, edge style={thick, blue, dotted}]
       \begin{deptext}[column sep=0.4cm]
          Let \& my \& people \& go! \\
       \end{deptext}
      \depedge[edge height=2ex]{4}{3}{nsubj}
      \depedge[edge height=2ex, label style={draw=black}, edge style={black, solid}]{1}{3}{dobj}
      \depedge[edge height=2ex, edge below, label style={draw=black}, edge style={black, solid}]{1}{4}{xcomp}
    \end{dependency}}
\end{gathered}
\end{equation}

\noindent{\bf Noun-modifying clauses:} Similarly, EUD links the empty subject of a finite relative clause to the corresponding argument of the external clause. We extend this behavior  to also cover \textbf{reduced relative clauses} (\ref{rrr-participle}a), and we follow \citet{candito-etal-2017-enhanced} in also including other relative clauses such as \textbf{noun-modifying participles} (\ref{rrr-participle}b).
\vspace{0.5em}
\begin{equation}a.\hspace{7em} b.\hspace{8em}
\label{rrr-participle}
\end{equation}
\scalebox{0.80}{
    \begin{dependency}[theme = default, label style={draw=blue}, edge style={thick, blue, dotted}]
       \begin{deptext}
          The \& neon \& god \& they \& made \hspace{1em} 
          \& A \& vision \& softly \& creeping \\
       \end{deptext}
       \depedge[edge height=2ex]{5}{3}{dobj}
       \depedge[edge height=2ex]{9}{7}{nsubj}
    \end{dependency}}

\noindent{\bf Adverbial clauses and ``dep'':} Adverbial modifier clauses that miss a subject, often modify the subject of the main clause. We propagate the external subject to be the subject of the internal clause.\footnote{In external clauses that include a subject and an object, ambiguity may arise as to which is to be modified.  We propagate both  and mark the edges as alternates (\textsc{Alt}, (\S\ref{sec:unc})).}
\vspace{0.5em}\begin{equation} \label{advcl}
\begin{gathered}
\scalebox{0.8}{
    \begin{dependency}[theme = default, label style={draw=blue}, edge style={thick, blue, dotted}]
       \begin{deptext}
          You \& shouldn't \& text \& while \& driving \\
       \end{deptext}
       \depedge[edge height=2ex]{5}{1}{nsubj}
    \end{dependency}}
\end{gathered}
\end{equation}

We observe  that many \textbf{dep} edges empirically behave like adverbial clauses, and  treat them similarly. We mark these edges as ``uncertain''.

\subsection{Parallel structures}
\label{sec:parallel}
The second type of conversions identifies parallel structures in which the latter  instance is elliptical, and share the missing core argument contributed by the former instance.

\noindent{\bf Apposition:} Similarly to the  PropS proposal \cite{stanovsky2016getting}, we share relations across {\em apposition parts}, making the two, currently hierarchical, phrase, more duplicate-like.\vspace{0.5em}
\begin{equation}
\begin{gathered}
\scalebox{0.8}{
    \begin{dependency}[theme = default, label style={draw=blue}, edge style={thick, blue, dotted}]
       \begin{deptext}
          E.T., \& the \& Extraterrestrial, \& phones \& home \\
       \end{deptext}
       \depedge[edge height=2ex]{4}{3}{nsubj}
       \depedge[edge below, edge height=2.5ex, label style={draw=black}, edge style={black, solid}]{4}{1}{nsubj}
       \depedge[edge height=2ex, label style={draw=black}, edge style={black, solid}]{1}{3}{appos}
    \end{dependency}}
\end{gathered}
\end{equation}

\noindent{\bf Modifiers in conjunction:} In modified coordinated constructions, we share prepositional (\ref{mogly}) and possessive (\ref{parents}) modifiers between the coordinated parts.  Since dependency trees are {\em inherently} ambiguous between conjoined modification and single-conjunt modification, (e.g, compare (\ref{mogly}) to ``Mogly was lost and raised by wolves", or (\ref{parents}) to ``my Father and E.T."), we mark both as \textsc{Unc}.

\begin{equation}
\begin{gathered}
\scalebox{0.8}{
    \begin{dependency}[theme = default, label style={draw=blue}, edge style={thick, blue, dotted}]
       \begin{deptext}[column sep=0.2cm]
          I \& was \& taught \& and \& raised \& by \& wolves \\
       \end{deptext}
       \depedge[edge height=2ex]{5}{7}{nmod(UNC)}
      \depedge[edge below, edge height=2ex, label style={draw=black}, edge style={black, solid}]{3}{7}{nmod}
    \end{dependency}}
\end{gathered}
\label{mogly}
\end{equation}

\begin{equation}
\begin{gathered}
\scalebox{0.8}{
    \begin{dependency}[theme = default, label style={draw=blue}, edge style={thick, blue, dotted}]
       \begin{deptext}[column sep=0.4cm]
          My \& father \& and \& mother \& met \& here \\
       \end{deptext}
       \depedge[edge height=2ex]{4}{1}{nmod:poss(UNC)}
       \depedge[edge below, edge height=3ex, label style={draw=black}, edge style={black, solid}]{2}{1}{nmod:poss}
    \end{dependency}}
\end{gathered}
\label{parents}
\end{equation}

\noindent{\bf Elaboration/Specification Clauses:} For noun nominal modifiers that have the form of an {\em elaboration} or {\em specification}, we share the head of the modified noun with its dependent modifier. That is, if the modification is marked by \emph{like} or \emph{such as} prepositions, we propagate the head noun to the nominal dependent.\vspace{0.5em}
\begin{equation}
\label{elaboration}
\begin{gathered}
\scalebox{0.8}{
    \begin{dependency}[theme = default, label style={draw=blue}, edge style={thick, blue, dotted}]
       \begin{deptext}[column sep=0.2cm]
          I \& enjoy \& fruits \& such \& as \& apples \\
       \end{deptext}
       \depedge[edge height=2ex]{2}{6}{dobj}
       \depedge[edge height=2ex, edge below, label style={draw=black}, edge style={black, solid}]{2}{3}{dobj}
    \end{dependency}}
\end{gathered}
\end{equation}

\noindent{\bf Indexicals:} the interpretation of locative and temporal indexicals such as \emph{here}, \emph{there} and \emph{now} depends on the situation and the speaker, and often modify not only the predicate but the entire situation. We therefore share the adverbial modification from the noun to the main verb. Due to their situation-specific nature, we mark these as \textsc{Unc}.\vspace{0.5em}
\begin{equation}
\begin{gathered}
\scalebox{0.8}{
    \begin{dependency}[theme = default, label style={draw=blue}, edge style={thick, blue, dotted}]
       \begin{deptext}[column sep=0.3cm]
          He \& wonders \& in \& these \& woods \& here \\
       \end{deptext}
       \depedge[edge height=2ex]{2}{6}{advmod(UNC)}
       \depedge[edge below, edge height=2ex, label style={draw=black}, edge style={black, solid}]{2}{5}{nmod}
       \depedge[edge below, edge height=2ex, label style={draw=black}, edge style={black, solid}]{5}{6}{advmod}
    \end{dependency}}
\end{gathered}
\end{equation}

\noindent{\bf Compounds:} \citet{shwartz-waterson-2018-olive} show that in many cases, compounds can be seen as having a multiple-head. Therefore, we share the existing relations across the compound parts.\vspace{0.5em}
\begin{equation}
\begin{gathered}
\scalebox{0.8}{
    \begin{dependency}[theme = default, label style={draw=blue}, edge style={thick, blue, dotted}]
       \begin{deptext}[column sep=0.4cm]
          I \& used \& canola \&  oil \\
       \end{deptext}
       \depedge[edge height=2ex]{2}{3}{dobj(UNC)}
       \depedge[edge below,edge height=2ex, label style={draw=black}, edge style={black, solid}]{2}{4}{dobj}
    \end{dependency}}
\end{gathered}
\end{equation}

\noindent As many compounds \emph{do} have a clear head (e.g. \emph{I used baby oil}, where \emph{baby} is clearly not the head), we mark these as uncertain.

\subsection{Syntactic Alternations}
\label{sec:alternation}
This type of conversions aim to unify syntactic variability. We identify structures that are syntactically different but share (some) semantic structure, and add arcs or nodes to expose the similarity.

\noindent{\bf The Passivization Alternation:}
Following \citet{candito-etal-2017-enhanced} we relate the \emph{passive} alteration to its \emph{active} variant.\vspace{0.5em}
\begin{equation}
\begin{gathered}
\scalebox{0.8}{
    \begin{dependency}[theme = default, label style={draw=blue}, edge style={thick, blue, dotted}]
       \begin{deptext}
          The \& Sheriff \& was \& shot \& by \& Bob \\
       \end{deptext}
       \depedge[edge height=2ex]{4}{6}{nsubj}
       \depedge[edge height=2ex]{4}{2}{dobj}
       \depedge[edge below, edge height=2ex, label style={draw=black}, edge style={black, solid}]{4}{2}{nsubjpass}
       \depedge[edge below, edge height=2ex, label style={draw=black}, edge style={black, solid}]{4}{6}{nmod:by}
    \end{dependency}}
\end{gathered}
\end{equation}

\noindent{\bf Hyphen reconstruction:} Noun-verb Hyphen Constructions (HC) which are modifying a nominal can be seen as conveying the same information as a copular sentence wherein the noun is the subject and the verb is the predicate. To explicitly indicate this, 
we add to all modifying noun-verb HCs  a {\em subject} and a {\em modifier} relation originating at  the verb-part of the HC.\vspace{0.5em}
\begin{equation}
\begin{gathered}
\scalebox{0.8}{
    \begin{dependency}[theme = default, label style={draw=blue}, edge style={thick, blue, dotted}]
       \begin{deptext}[column sep=0.4cm]
          A \& Miami \& - \& based \& company \\
       \end{deptext}
       \depedge[edge height=2ex]{4}{5}{nsubj}
       \depedge[edge height=2ex]{4}{2}{nmod}
       \depedge[edge below, edge height=2ex, label style={draw=black}, edge style={black, solid}]{5}{4}{amod}
       \depedge[edge below, edge height=2ex, label style={draw=black}, edge style={black, solid}]{4}{2}{compound}
    \end{dependency}}
\end{gathered}
\end{equation}

\noindent{\bf Adjectival modifiers:}
Adjectival modification can be viewed as capturing the same information as a predicative copular sentence conveying the same meaning (so, ``a green apple'' implies that ``an apple is green''). To explicitly capture this productive implication, we add a subject relation from each adjectival modifier to its corresponding modified noun.\vspace{0.5em}
\begin{equation}
\begin{gathered}
\scalebox{0.8}{
    \begin{dependency}[theme = default, label style={draw=blue}, edge style={thick, blue, dotted}]
       \begin{deptext}[column sep=0.2cm]
          I \& see \& dead \& people \\
       \end{deptext}
       \depedge[edge height=2ex]{3}{4}{nsubj}
    \end{dependency}}
\end{gathered}
\end{equation}

\noindent{\bf Genitive Constructions:} Genitive cases can be alternatively expressed as a compound. We add a compound relation to unify the expression of genitives across {\em X of Y} and {\em compound} structures.
\begin{equation}
\begin{gathered}
\scalebox{0.8}{
    \begin{dependency}[theme = default, label style={draw=blue}, edge style={thick, blue, dotted}]
       \begin{deptext}[column sep=0.2cm]
          Army \& of \& zombies \\
       \end{deptext}
       \depedge[edge height=2ex]{1}{3}{compound}
    \end{dependency}}
\end{gathered}
\end{equation}

\subsection{\bf Event-Centered Representations}\vspace{-0.5em}\label{sec:state}
In many sentences, the finite root predicate does not indicate the main event. Instead, a verb in the subordinated clause
expresses the event, and the finite verb acts as its modifier.
For example, in sentences like ``He started working", ``He seems to work there", the main event indicated is ``work", while the root predicates (``started'', ``seemed'') modify this event.
Here, we present a chain of changes that puts emphasis on {\em events} by delegating copular and tense auxiliaries (is, was), evidentials (seem, say) and various aspectual verbs (started, continued) to be clausal modifiers, rather than heads of the sentence.
This creates a further challenge, since there is a prevalent discrepancy between predicative sentences such as ``He works" and copular sentences  as ``He is smart". The UD structure for the latter lacks a node that  clearly indicates a  {\em stative} event (in \citet{vendler}'s terminology). We remedy this by adding a node to represent the STATE and have tense, aspect, modality and evidentiality directly  modifying it.\footnote{Pragmatically, some users prefer to not have non-word nodes. pyBART supports this by providing a mode that treats the copula as the head, retaining the other modifications.}

\noindent{\bf Copular Sentences and Stative Predicates:}
We added to all copula constructions new node named \emph{STATE}, which represents  the {\em stative} event introduced by the copular clause. This node becomes the root, and we rewire the entire clause around this \emph{STATE}. By doing so we unify it with the structures of  clauses with finite predicative. Once we added the \emph{STATE} node, we form a new relation, termed \emph{ev}, to mark event/state modifications.
The resulting structure is as follows:\vspace{0.5em} 
\begin{equation}
\begin{gathered}
\scalebox{0.8}{
    \begin{dependency}[theme = default, label style={draw=blue}, edge style={thick, blue, dotted}]
       \begin{deptext}[column sep=0.2cm]
          Tomorrow \& is \& STATE \& another \& day \& \\
       \end{deptext}
       \depedge[edge height=4ex]{3}{1}{nsubj}
       \depedge[edge height=2ex]{3}{2}{ev}
       \depedge[edge height=2ex]{3}{5}{xcomp}
       \depedge[edge below, edge height=4ex, label style={draw=red}, edge style={red,densely dashed}]{5}{1}{nsubj}
       \depedge[edge below, edge height=2ex, label style={draw=red}, edge style={red,densely dashed}]{5}{2}{cop}
    \end{dependency}}
\end{gathered}
\end{equation}

\noindent{\bf Evidential  reconstructions:}
We can now explicitly mark properties of events as dependents of the verbal or stative root by means of the label {\em ev}. We do so, using verbs' white-lists, for verbs  marking evidentiality (\ref{state-ev}) and  for reported-speech (\ref{state-rep}).
\begin{equation}
\begin{gathered}
\scalebox{0.85}{
\vspace{-0.5em}
    \begin{dependency}[theme = default, label style={draw=blue}, edge style={thick, blue, dotted}]
       \begin{deptext}
          Sam \& seems \& to \& like \& them. \& \& They \& seem \& STATE \& nice. \\
       \end{deptext}
       \depedge[edge height=4ex]{4}{1}{nsubj}
       \depedge[edge height=2ex]{4}{2}{ev}
       
       \depedge[edge height=4ex]{9}{7}{nsubj}
       \depedge[edge height=2ex]{9}{8}{ev}
       \depedge[edge height=2ex]{9}{10}{xcomp}
       \depedge[edge below, edge height=2ex, label style={draw=red}, edge style={red,densely dashed}]{2}{1}{nsubj}
       \depedge[edge below, edge height=2ex, label style={draw=red}, edge style={red,densely dashed}]{2}{4}{xcomp}
       \depedge[edge below, edge height=2ex, label style={draw=red}, edge style={red,densely dashed}]{8}{7}{nsubj}
       \depedge[edge below, edge height=2ex, label style={draw=red}, edge style={red,densely dashed}]{8}{10}{xcomp}
    \end{dependency}}
\end{gathered}
\label{state-ev}
\end{equation}
\begin{equation}
\begin{gathered}
\scalebox{0.85}{
    \begin{dependency}[theme = default, label style={draw=blue}, edge style={thick, blue, dotted}]
       \begin{deptext}[column sep=-0.05cm]
          The \& Media \& reported \& that \& peace \& was \& achieved \\
       \end{deptext}
       \depedge[edge height=2ex]{7}{3}{ev}
       \depedge[edge below, edge style={black, solid}, edge height=2ex, label style={draw=black}]{3}{7}{ccomp}
    \end{dependency}}
\end{gathered}
\label{state-rep}
\end{equation}

\noindent{\bf Aspectual constructions:}
Finally, we can now also mark aspectual verbs as modifying the complement (matrix) verb denoting the main event. The  complement (matrix) verb becomes the root of the dependency structure, and we add the new \emph{ev} relation to mark the  aspectual modification of the event.
\vspace{0.5em}\begin{equation}
\begin{gathered}
\scalebox{0.8}{
    \begin{dependency}[theme = default, label style={draw=blue}, edge style={thick, blue, dotted}]
       \begin{deptext}[column sep=0.4cm]
          He \& started \& talking \& funny \\
       \end{deptext}
       \depedge[edge height=4ex]{3}{1}{nsubj}
       \depedge[edge height=2ex]{3}{2}{ev}
       \depedge[edge below, edge height=2ex, label style={draw=red}, edge style={red,densely dashed}]{2}{1}{nsubj}
       \depedge[edge below, edge height=2ex, label style={draw=red}, edge style={red,densely dashed}]{2}{3}{xcomp}
    \end{dependency}}
\end{gathered}
\end{equation}

\section{Evaluation}
\label{sec:eval}
Our proposed representation attempts to target information-seeking applications, but is it effective? We evaluate the resulting graph structures against the UD and Enhanced UD representations, in the context of a relation-extraction (RE) task. 
Concretely, we evaluate the representations on their ability to perform pattern-based RE on the TACRED dataset \cite{zhang2017tacred}.

We use an automated and reproducible methodology: for each of the representations, we use the RE train-set to acquire extraction patterns. We then apply the patterns to the dev-set, compute F1-scores, and, for each relation, filter the patterns that hurt F1-score. We then apply the filtered pattern-set to the test-set, and report F1 scores.

To acquire extraction patterns, we use the following procedure: given a labeled sentence consisting of a relation name and the sentence indices of the two entities participating in the relation, we compute the shortest dependency path between the entities, ignoring edge directions. We then form an extraction pattern from the directed edges on this path. We consult a list of trigger words \cite{yu-etal-2015-read} collected for the different relations. If a trigger word or its lemma is found on the path, we form an unlexicalized path except for the trigger word (i.e. {\small\textsf{E1 $<$nsubj ``founded'' $>$dobj $>$compound E2}}).
If no trigger-word is found, the path is lexicalized with the word's lemmas (i.e. {\small\textsf{E1 $<$nsubj ``reduce'' $>$dobj ``activity'' $>$compound E2}}).

\begin{table}[t]
\centering
\scalebox{0.8}{
\begin{tabular}{||l c c c||} 
 \hline
 Representation & Precision & Recall & F1 \\ [0.5ex] 
 \hline\hline
 UD & 76.53 & 30.65 & 43.77 \\ 
 Enhanced UD & 77.63 & 32.37 & 45.69 \\
 Ours(w/o-Enhanced) & 73.96 & 33.48 & 46.09 \\
 Ours & 74.62 & 36.65 & {\bf 49.15} \\ [1ex] 
 \hline
\end{tabular}}
\caption{Effectiveness of the different representations on the TACRED relation extraction task.}
\label{table:1}
\end{table}

We use this procedure to compare UD, Enhanced UD (EUD), 
BART without EUD enhancements, and full BART, which is a superset of Enhanced UD (Table \ref{table:1}). BART achieves a substantially higher F1 score of 49.15\%, an increase of 5.5 F1 points over UD, and $3.5$ F1 points above Enhanced UD. It does so by substantially improving recall while somewhat decreasing precision.

\begin{figure}
    \includegraphics[width=\linewidth]{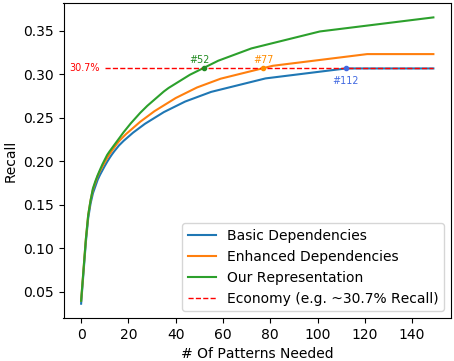}
    \caption{Economy comparison: Recall vs number of patterns, for the different representations.} 
    \label{fig:1}
\end{figure}

We also  consider \emph{economy}: the number of different patterns needed to achieve a given recall level.
Figure \ref{fig:1} plots the achieved recall against the number of patterns.
As the curves show, Enhanced UD is more economic than UD, and our representation is substantially more economic than both.
To achieve 30.7\% recall (the maximal recall of UD), UD requires 112 patterns, EUD requires 77 patterns, while \rep needed only 52 patterns.

\section{Conclusion}
We propose a syntax-based representation that aims to make the event structure and as many lexical relations as possible explicit, for the benefit of downstream information-seeking applications. We provide a Python API that converts UD trees to this representation, and  demonstrate its empirical benefits  on a relation extraction task.
\section*{Acknowledgements}
This project has  received funding from the European Research Council (ERC) under the European Union's Horizon2020 research and innovation programme, grant agreement   802774 (iEXTRACT) and grant agreement   677352 (NLPRO).

\bibliography{pyBART}
\bibliographystyle{acl_natbib}

\newpage
\appendix
\section{Appendix}

\subsection{Additional examples of BART structures}
\label{sec:complementaryExamples}
The following are additional examples that we could not fit into the space constraints of the paper.

\paragraph{Complement control}
Example (\ref{complement-control1}) shows an example of linking the external subject to a controlled finite verb.
The following complementary example shows linking the subject to a controlled gerund:
\begin{equation}
\begin{gathered}
\scalebox{0.8}{
    \begin{dependency}[theme = default, label style={draw=blue}, edge style={thick, blue, dotted}]
       \begin{deptext}[column sep=0.4cm]
          I \& decided \& going \& home \\
       \end{deptext}
      \depedge[edge height=2ex, edge below]{3}{1}{nsubj}
      \depedge[edge height=2ex, label style={draw=black}, edge style={black, solid}]{2}{3}{xcomp}
      \depedge[edge height=2ex, label style={draw=black}, edge style={black, solid}]{2}{1}{nsubj}
      \depedge[edge height=2ex, label style={draw=black}, edge style={black, solid}]{3}{4}{advmod}
    \end{dependency}}
\end{gathered}
\end{equation}

\paragraph{Elaboration/Specification Clauses} 
Example (\ref{elaboration}) shows a specification clause connected as an object to the root. The following is a complementary example of using the \emph{like} elaboration-preposition, in which the modifier noun is a \emph{subject} dependant of its head. The modified noun inherits the \emph{subject} relation from its modifier head.

\begin{equation}
\begin{gathered}
\scalebox{0.8}{
    \begin{dependency}[theme = default, label style={draw=blue}, edge style={thick, blue, dotted}]
       \begin{deptext}[column sep=0.4cm]
          People \& like \& you \& should \& feel \& lucky \\
       \end{deptext}
      \depedge[edge height=2ex, edge below]{5}{3}{nsubj}
      \depedge[edge height=2ex, label style={draw=black}, edge style={black, solid}]{5}{6}{xcomp}
      \depedge[edge height=7ex, label style={draw=black}, edge style={black, solid}]{5}{1}{nsubj}
      \depedge[edge height=2ex, label style={draw=black}, edge style={black, solid}]{3}{2}{case}
      \depedge[edge height=4.5ex, label style={draw=black}, edge style={black, solid}]{1}{3}{nmod}
      \depedge[edge height=2ex, label style={draw=black}, edge style={black, solid}]{5}{4}{aux}
    \end{dependency}}
\end{gathered}
\end{equation}

\paragraph{Adjectival modifier}

\begin{equation}
\begin{gathered}
\scalebox{0.8}{
    \begin{dependency}[theme = default, label style={draw=blue}, edge style={thick, blue, dotted}]
       \begin{deptext}[column sep=0.4cm]
          The \& smart \& one \& waited \& patently \\
       \end{deptext}
      \depedge[edge height=2ex, edge below]{2}{3}{nsubj}
      \depedge[edge height=2ex, label style={draw=black}, edge style={black, solid}]{4}{3}{nsubj}
      \depedge[edge height=4.5ex, label style={draw=black}, edge style={black, solid}]{3}{1}{det}
      \depedge[edge height=2ex, label style={draw=black}, edge style={black, solid}]{3}{2}{amod}
      \depedge[edge height=2ex, label style={draw=black}, edge style={black, solid}]{4}{5}{advmod}
    \end{dependency}}
\end{gathered}
\end{equation}

\newpage
\paragraph{Copular Sentences and Stative Predicates}  
We show additional examples of these transformations, with explicit comparison to UD.

\begin{equation}UD:\hspace{4em}BART:\hspace{7em}
\end{equation}
\scalebox{0.8}{
    \begin{dependency}[theme = default, label style={draw=blue}, edge style={thick, blue, dotted}]
       \begin{deptext}
          Sally \& is \& smart \hspace{1em} \& Sally \& is \& STATE \& smart \\
       \end{deptext}
      \depedge[edge height=6ex, label style={draw=red}, edge style={red,densely dashed}]{3}{1}{nsubj}
      \depedge[edge height=3ex, label style={draw=red}, edge style={red,densely dashed}]{3}{2}{cop}
      
      \depedge[edge height=3ex]{6}{7}{xcomp}
      \depedge[edge height=3ex]{6}{5}{ev}
      \depedge[edge height=6ex]{6}{4}{nsubj}
      \depedge[edge height=2ex, edge below]{4}{7}{amod}
    \end{dependency}}

\begin{equation}UD:\hspace{4em}BART:\hspace{7em}
\end{equation}
\scalebox{0.8}{
    \begin{dependency}[theme = default, label style={draw=blue}, edge style={thick, blue, dotted}]
       \begin{deptext}
          He \& is \& the \& man \hspace{1em} \& He \& is \& STATE \& the \& man \\
       \end{deptext}
      \depedge[edge height=9ex, label style={draw=red}, edge style={red,densely dashed}]{4}{1}{nsubj}
      \depedge[edge height=6ex, label style={draw=red}, edge style={red,densely dashed}]{4}{2}{cop}
      \depedge[edge height=3ex, label style={draw=black}, edge style={black, solid}]{4}{3}{det}
      
      \depedge[edge height=3ex, label style={draw=black}, edge style={black, solid}]{9}{8}{det}
      \depedge[edge height=6ex]{7}{9}{xcomp}
      \depedge[edge height=3ex]{7}{6}{ev}
      \depedge[edge height=6ex]{7}{5}{nsubj}
    \end{dependency}}

\begin{equation}UD:\hspace{4em}BART:\hspace{7em}
\end{equation}
\scalebox{0.8}{
    \begin{dependency}[theme = default, label style={draw=blue}, edge style={thick, blue, dotted}]
       \begin{deptext}
          They \& are \& from \& Israel \hspace{1em} \& They \& are \& STATE \& from \& Israel \\
       \end{deptext}
      \depedge[edge height=9ex, label style={draw=red}, edge style={red,densely dashed}]{4}{1}{nsubj}
      \depedge[edge height=6ex, label style={draw=red}, edge style={red,densely dashed}]{4}{2}{cop}
      \depedge[edge height=3ex, label style={draw=black}, edge style={black, solid}]{4}{3}{case}
      
      \depedge[edge height=3ex, label style={draw=black}, edge style={black, solid}]{9}{8}{case}
      \depedge[edge height=6ex]{7}{9}{nmod}
      \depedge[edge height=3ex]{7}{6}{ev}
      \depedge[edge height=6ex]{7}{5}{nsubj}
    \end{dependency}}

\begin{equation}UD:\hspace{4em}BART:\hspace{7em}
\end{equation}
\scalebox{0.8}{
    \begin{dependency}[theme = default, label style={draw=blue}, edge style={thick, blue, dotted}]
       \begin{deptext}
          Sam \& is \& to \& be \& a \& man \hspace{1em} \& Sam \& is \& to \& be \& STATE \& a \& man \\
       \end{deptext}
      \depedge[edge height=3ex, label style={draw=red}, edge style={red,densely dashed}]{2}{1}{nsubj}
      \depedge[edge height=12ex, label style={draw=red}, edge style={red,densely dashed}]{2}{6}{xcomp}
      \depedge[edge height=9ex, label style={draw=red}, edge style={red,densely dashed}]{6}{3}{mark}
      \depedge[edge height=6ex, label style={draw=red}, edge style={red,densely dashed}]{6}{4}{cop}
      \depedge[edge height=3ex, label style={draw=black}, edge style={black, solid}]{6}{5}{det}
      
      \depedge[edge height=3ex, label style={draw=black}, edge style={black, solid}]{13}{12}{det}
      \depedge[edge height=6ex]{11}{13}{xcomp}
      \depedge[edge height=3ex]{11}{10}{ev}
      \depedge[edge height=3ex]{10}{8}{ev}
      \depedge[edge below, edge height=3ex]{10}{9}{mark}
      \depedge[edge height=6ex]{11}{7}{nsubj}
    \end{dependency}}

\begin{equation}UD:\hspace{4em}BART:\hspace{7em}
\end{equation}
\scalebox{0.8}{
    \begin{dependency}[theme = default, label style={draw=blue}, edge style={thick, blue, dotted}]
       \begin{deptext}
          Sam \& sounds \& funny \hspace{1em} \& Sam \& sounds \& STATE \& funny \\
       \end{deptext}
      \depedge[edge height=3ex, label style={draw=red}, edge style={red,densely dashed}]{2}{1}{nsubj}
      \depedge[edge height=3ex, label style={draw=red}, edge style={red,densely dashed}]{2}{3}{xcomp}
      
      \depedge[edge height=3ex]{6}{7}{xcomp}
      \depedge[edge height=3ex]{6}{5}{ev}
      \depedge[edge height=6ex]{6}{4}{nsubj}
      \depedge[edge height=2ex, edge below]{4}{7}{amod}
    \end{dependency}}
    
\begin{equation}UD:\hspace{4em}BART:\hspace{7em}
\end{equation}
\scalebox{0.8}{
    \begin{dependency}[theme = default, label style={draw=blue}, edge style={thick, blue, dotted}]
       \begin{deptext}
          Sam \& seems \& happy \hspace{1em} \& Sam \& seems \& STATE \& happy \\
       \end{deptext}
      \depedge[edge height=3ex, label style={draw=red}, edge style={red,densely dashed}]{2}{1}{nsubj}
      \depedge[edge height=3ex, label style={draw=red}, edge style={red,densely dashed}]{2}{3}{xcomp}
      
      \depedge[edge height=3ex]{6}{7}{xcomp}
      \depedge[edge height=3ex]{6}{5}{ev}
      \depedge[edge height=6ex]{6}{4}{nsubj}
      \depedge[edge height=2ex, edge below]{4}{7}{amod}
    \end{dependency}}
    
\begin{equation}UD:\hspace{4em}BART:\hspace{7em}
\end{equation}
\scalebox{0.8}{
    \begin{dependency}[theme = default, label style={draw=blue}, edge style={thick, blue, dotted}]
       \begin{deptext}
          Sam \& seems \& to \& like \& them  \& \hspace{1em} \& Sam \& seems \& to \& like \& them \\
       \end{deptext}
      \depedge[edge height=3ex, label style={draw=red}, edge style={red,densely dashed}]{2}{1}{nsubj}
      \depedge[edge height=3ex, label style={draw=red}, edge style={red,densely dashed}]{2}{4}{xcomp}
      \depedge[edge height=3ex, label style={draw=red}, edge style={red,densely dashed}]{4}{5}{dobj}
      
      \depedge[edge height=3ex, edge style={black, solid}]{10}{11}{dobj}
      \depedge[edge height=3ex]{10}{8}{ev}
      \depedge[edge height=6ex]{10}{7}{nsubj}
    \end{dependency}}

\begin{equation}UD:\hspace{4em}BART:\hspace{7em}
\end{equation}
\scalebox{0.8}{
    \begin{dependency}[theme = default, label style={draw=blue}, edge style={thick, blue, dotted}]
       \begin{deptext}[column sep=-0.1em]
          Sally \& began \& walking \& home \hspace{1em} \& Sally \& began \& walking \& home \\
       \end{deptext}
      \depedge[edge height=3ex, label style={draw=red}, edge style={red,densely dashed}]{2}{1}{nsubj}
      \depedge[edge height=3ex, label style={draw=red}, edge style={red,densely dashed}]{2}{3}{xcomp}
      \depedge[edge height=3ex, label style={draw=black}, edge style={black, solid}]{3}{4}{advmod}
      
      \depedge[edge height=3ex, label style={draw=black}, edge style={black, solid}]{7}{8}{advmod}
      \depedge[edge height=3ex]{7}{6}{ev}
      \depedge[edge height=6ex]{7}{5}{nsubj}
    \end{dependency}}

\end{document}